\documentclass{l4dc2027}

\title[Drifting, Discriminating, and Directioning for Action Coordination]{Drifting, Discriminating, and Directioning for Action Coordination in Higher-dimensional Action Space}
\usepackage{times}

\author{%
 \Name{Taisuke Kobayashi} \Email{kobayashi@nii.ac.jp}\\
 \addr National Institute of Informatics and The Graduate University for Advanced Studies, SOKENDAI
}

\begin{document}

\maketitle
\thispagestyle{empty}

\begin{abstract}%
This paper proposes a lightweight policy model to mitigate the difficulty of handling higher-dimensional action spaces in reinforcement learning (RL), alongside an algorithm to effectively train it.
To address this, we focus on drifting models, which can implicitly represent complex distributions and infer actions in a single forward pass.
However, while drifting models can be straightforwardly trained via backpropagation of the Q-gradient, this alone cannot be expected to appropriately capture the desired action coordination.
Therefore, this paper constructs a three-step learning algorithm to encourage the action coordination.
First, the source distribution of the drifting model is co-optimized as a digonal policy model.
This conditions the drifting model to attach the coordination, which cannot be expressed by a diagonal distribution, as residuals.
Second, a discriminator model before and after drifting is introduced with adversarial training for a regularization term.
This actively promotes the extraction of the coordination that fundamentally separates the two distributions.
Third, best-of-N sampling is applied for action selection during rollouts in order to give directionality toward/beyond the regions where the coordination is insufficiently extracted/sufficiently discovered.
The proposed method consistently improves the sample efficiency and the final control performance over the conventional diagonal distribution modeling across three locomotion tasks with high-dimensional action spaces.
Furthermore, the analysis indicates that this improvement is achieved by reducing variance in certain action dimensions, effectively coupling them with other actions as action coordination.
\end{abstract}


\section{Introduction}

Reinforcement learning (RL) \citep{sutton2018reinforcement} is a promising methodology for learning control policies from experience, and recent advances in robot control have increasingly relied on RL to perform a wide range of tasks \citep{rudin2022learning,radosavovic2024real}.
However, it is well known that as the number of joints increases like humanoid robots and the action space expands accordingly, the difficulty of learning increases.
Currently, this challenge is often circumvented by limiting the exploration space using heuristics based on domain knowledge.
For example, this involves designing rich and dense reward functions to finely guide the robot toward optimal actions \citep{van2024revisiting,liao2026beyondmimic}, or focusing on the modularity of the robot's physical structure to divide the control policy into multiple parts and handle subspaces \citep{liu2024masq,zhang2026falcon}.
Thanks to the efforts of many engineers, these workarounds function well.
However, fundamentally, restricting the exploration space risks hindering the discovery of global optimal solutions.
Ideally, we need end-to-end learning algorithms that can learn effectively even in high-dimensional action spaces without relying on such domain knowledge.

This paper hypothesizes that the difficulty in high-dimensional action spaces stems from the fact that policies in major algorithms/libraries are typically modeled using ``diagonal'' probability distributions \citep{chou2017improving,kobayashi2019student,schwarke2025rsl,kim2026flashsac}.
In other words, as the action space grows larger, it should be rare for all action axes to be independent; however, diagonal distributions assume this independence, leading to a discrepancy.
This likely results in increased unnecessary exploration and interference between actions, which degrades sample efficiency and ultimately prevents the discovery of a solution.
A naive solution would be to adopt a non-diagonal multivariate probability distribution model like multivariate Gaussian.
However, since dependencies between actions are expected to be state-dependent, this approach requires predicting $|\mathcal{A}|(|\mathcal{A}|-1)/2$ covariance parameters for each state corresponding to the action dimension $|\mathcal{A}|$, resulting in a lack of scalability.

Recent advances in generative models have led to the emergence of diffusion policies and flow-matching policies, which can implicitly model highly expressive probability distributions \citep{chi2025diffusion,zhang2025flowpolicy}.
Since these approaches reflect the dependencies between actions during the denoising process, they have the potential to address challenges in high-dimensional action spaces.
However, the iterative computations required to generate actions are computationally expensive, making them not suited for tasks such as locomotion control, where real-time control is often critical.
Furthermore, since these approaches involve reformulating optimization problems based on vector fields \citep{ren2025diffusion,mcallister2026flow}, they have low compatibility with traditional RL algorithms and cannot be scaled up at low cost.

Therefore, this paper focuses on drifting models \citep{deng2026drifting} among these new probabilistic modeling approaches.
By embedding iterative computations in the generative process within a feedforward network, this approach enables the conversion of a noise source into a target random variable (in this paper, action) in a single forward pass.
Alternatively, it can be described as an extension of normalizing flows \citep{papamakarios2021normalizing}, the precursor to these techniques, that sacrifices invertibility and simplified likelihood calculations in favor of enhanced expressiveness.
This model avoids the issue of computational cost, and by treating it as a kind of normalizing flows, it can be easily combined with RL algorithms using Q-gradient \citep{haarnoja2018soft,fujimoto2018addressing}, not policy likelihood.

\begin{figure}[!t]
    \centering
    \includegraphics[keepaspectratio=true,width=0.96\linewidth]{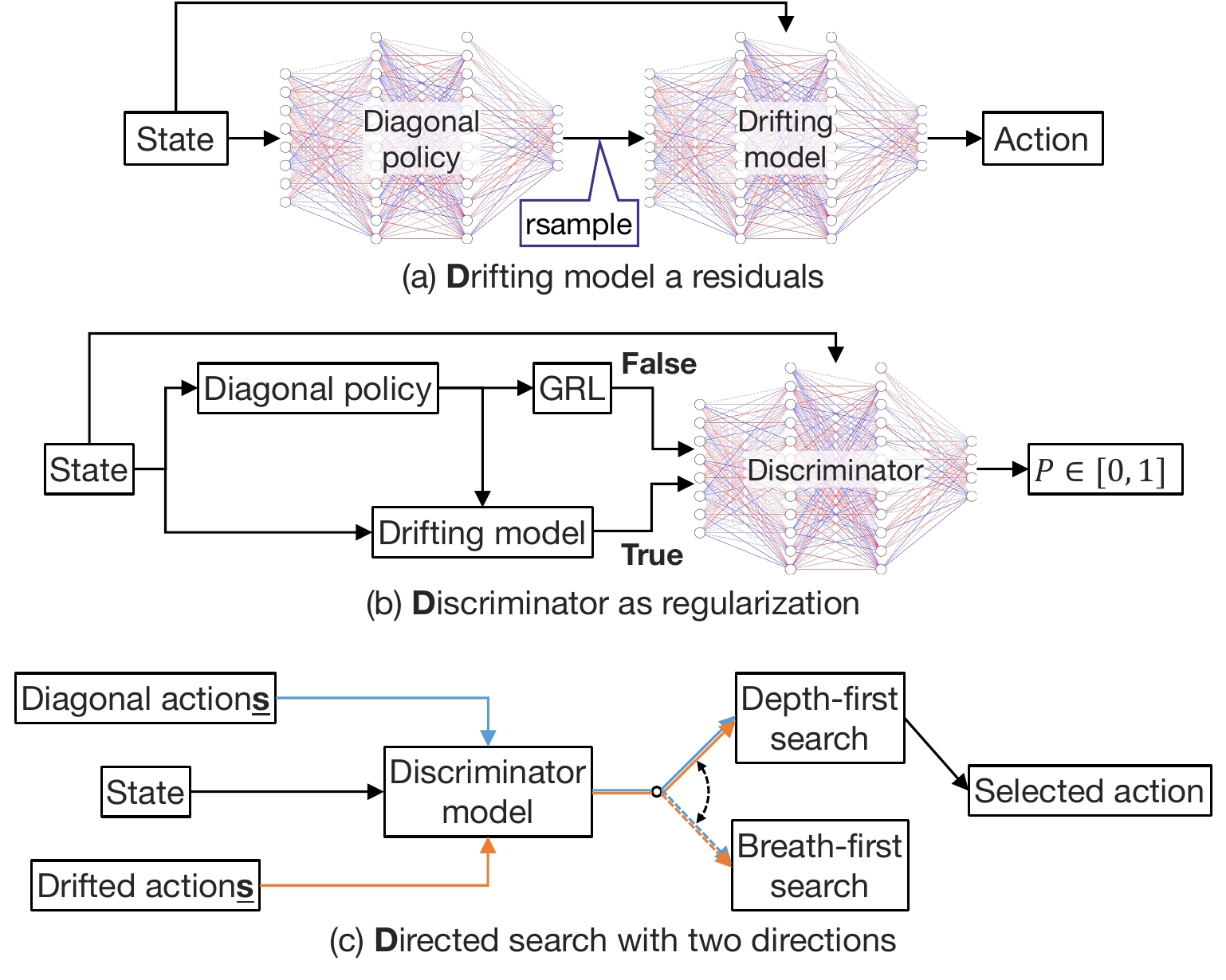}
    \caption{Three techniques for actively extracting action coordination}
    \label{fig:proposal}
\end{figure}

This paper proposes a new RL algorithm that utilizes these drifting models to handle action coordination.
However, simply optimizing the drifting models using the Q-gradient did not yield the expected performance.
Therefore, three techniques are introduced in sequence that enable the more proactive extraction of the action coordination (see Fig.~\ref{fig:proposal}).
The key idea is to encourage the drifting network model to extract only the coordination components.

First, the noise source in drifting models has a clear role.
Specifically, prior to drifting is defined as a diagonal policy, and the inter-independent action generated from it is made to drift.
By doing so, it is structurally expected that the drifting model contains only coordination components as residuals \citep{johannink2019residual,wen2022multi}.

Second, a new regularization term is designed to promote this role separation.
Specifically, a discriminator model, which distinguishes between action before and after drifting, is optimized by adversarial training \citep{goodfellow2020generative,kobayashi2024revisiting} in conjunction with RL.
By doing so, the coordination components are more actively extracted into the drifting model for appropriately identifying the drifted action.

Third, a directed search with the discriminator is conducted for rollouts.
Specifically, best-of-N sampling \citep{stiennon2020learning,kobayashi2026flexible} for maximizing mutual information represented by the discriminator \citep{letizia2021discriminative} explores areas where learning is insufficient as a breadth-first approach; or areas where the action coordination has been identified as a depth-first approach.
By doing so, high-value experience data is more efficiently collected than through simple probabilistic exploration.

This paper reports that, compared to the diagonal policy model, these modifications gradually improve learning performance (i.e. sample efficiency and final control performance) across three locomotion tasks in dm\_control \citep{tunyasuvunakool2020dm_control}.
Furthermore, by analyzing the learned policy, it is demonstrated that the actions generated by the proposed method reduce the variance in certain action dimensions.
In other words, we conclude that the proposed method reduces the number of effective action dimensions and fosters coordinated behavior.

\section{Preliminaries}

\subsection{Reinforcement learning with Q-gradient}

This paper focuses on the basic Markov decision process (MDP) as a problem setting in RL to simplify the discussion \citep{sutton2018reinforcement}.
Specifically, given that the state $s \in \mathcal{S} \subseteq \mathbb{R}^{|\mathcal{S}|}$ of an environment is determined probabilistically by the initial state distribution $p_0$ or the state transition probability distribution $p_e$, an agent intervenes through its action $a \in \mathcal{A} \subseteq \mathbb{R}^{|\mathcal{A}|}$ in the transition for getting the desired state and action, represented by the reward $r \in \mathcal{R} \subseteq \mathbb{R}$.
The agent possesses a policy $\pi$, a state-dependent probability distribution that generates $a$, with $\phi$ the network-model parameters, and optimizes $\phi$ to maximize the sum of rewards from any time $t$ into the future (referred to as the return) or its expectation, $\mathbb{E}_{p_e\pi} [\sum_{k=0}^\infty \gamma^k r_{t+k}]$, where $\gamma \in [0, 1)$ is the discount factor.

With this MDP, by approximating and predicting the expected value of the return as a function, we can easily evaluate the direction of optimization for $a$ and $\pi$.
When this is modeled as an action value function $Q$ with $\theta$ the network-model parameters, the temporal difference (TD) method described below is widely used for the optimization of $\theta$.
\begin{align}
    \theta^\ast = \arg \min_{\theta} \mathbb{E}_{(s_i, a_i, s^\prime_i, r_i) \sim D}\left[ (r_i + \gamma \mathrm{sg}(Q_\theta(s^\prime_i, a^\prime)) - Q_\theta(s_i, a_i) )^2 \right]
\end{align}
where, $D$ is a replay buffer that has accumulated the experience $(s_t, a_t, s_{t+1}=s^\prime_t, r_t)$ at $t$ up to its capacity $|D|$.
In addition, $\mathrm{sg}(\cdot)$ is a stop-gradient operation, and in this paper, $a^\prime$ is sampled from $\pi_\phi(\cdot \mid s^\prime_i)$ (i.e. expected SARSA is employed).
In other words, $r_i + \gamma \mathrm{sg}(Q_\theta(s^\prime_i, a^\prime))$ is the Q-target as the bootstrap method, and it is common practice to compute it using a target network with $\bar{\theta}$, where $\theta$ has been smoothed, to ensure computational stability.

If $Q$ is differentiable (by approximating it with neural networks), the gradient can be backpropagated to the input $a$.
Furthermore, by using the reparameterization trick \citep{kingma2014auto} to generate $a$ from $\pi$, the gradient can be propagated to $\pi$ (and thus to $\phi$).
Therefore, the agent can achieve the original objective of maximizing the expected return through the following minimization problem.
\begin{align}
    \phi^\ast = \arg \min_{\phi} \mathbb{E}_{(s_i) \sim D}\left[ - Q_\theta(s_i, a(s_i)) \right]
    \quad \mathrm{s.t.}\quad
    a(s) = \mathrm{rt}(\pi_\phi(\cdot \mid s))
\end{align}
where, $\mathrm{rt}(\cdot)$ denotes the reparameterization trick.

Note that the above is a naive algorithm, and this paper uses soft actor-critic \citep{haarnoja2018soft} and its modification \citep{han2021max,nauman2025decoupled} to improve learning performance and to introduce diversity into the noise sources of the drifting model.
Specifically, an entropy term for the policy is introduced to both the Q-target and Q-gradient.
However, while entropy is approximated by the negative log-likelihood, incorporating the drifting model, which will be introduced in the next section, makes this calculation difficult.
Instead, the entropy term of the source distribution is added to them in this study.
Even in this case, it is still possible to introduce diversity into the noise sources.

\subsection{Drifting model}

The drifting model is a new generative model with a single feedforward transformation \citep{deng2026drifting}.
Specifically, we have a mapping function, $f: \mathcal{C} \mapsto \mathcal{D}$, with $\mathcal{C}$ the space of noise source and $\mathcal{D}$ the target space.
When $\epsilon \in \mathcal{C}$ is generated as a random variable from probability distribution $p$, the transformed one, $x \in \mathcal{D}$, is also regarded as a random variable implicitly.
By decomposing $f$ into the combination of monotone functions, $m_i: \mathcal{C}_i \mapsto \mathcal{D}_i$, on subspaces ($i=1,\ldots,M$), the corresponding probability distribution $q$ can be given as follows:
\begin{align}
    q(x) = \sum_{i=1}^M p(\epsilon) |\det J_{m_i}^{-1}(\epsilon)|
    \quad \mathrm{s.t.}\quad
    x = f(\epsilon) = \sum_{i=1}^M m_i(\epsilon)
\end{align}
where, $J_{m_i}$ denotes the Jacobian of $m_i$ over $\epsilon$.
In practice, since the above decomposition is intractable, it is best to assume that this likelihood cannot be computed.
Note that the case where $\mathcal{C} = \mathcal{D}$ and $f = m$ corresponds to normalizing flows, and these often involve additional structural constraints that simplify the calculation of the Jacobian \citep{papamakarios2021normalizing}.

Anyway, if we model $f$ using a highly expressive neural network, we can implicitly represent any probability distribution and generate corresponding data with inter-axis dependencies.
Although the training of this drifting model is originally based on vector fields, since $f$ is differentiable, this paper considers training it directly using the Q-gradient.
This can be implemented only by modifying the policy modeling from the conventional RL algorithm with the Q-gradient above.
However, in practice, we could not naively achieve the expected result, where the drifting model appropriately captures action coordination to improve learning performance.
Therefore, starting in the next section, three additional (albeit simple) techniques are proposed to implement RL efficiently and stably with this approach.

\section{Proposal}

\subsection{Overview}

\begin{figure}[!t]
    \centering
    \includegraphics[keepaspectratio=true,width=0.96\linewidth]{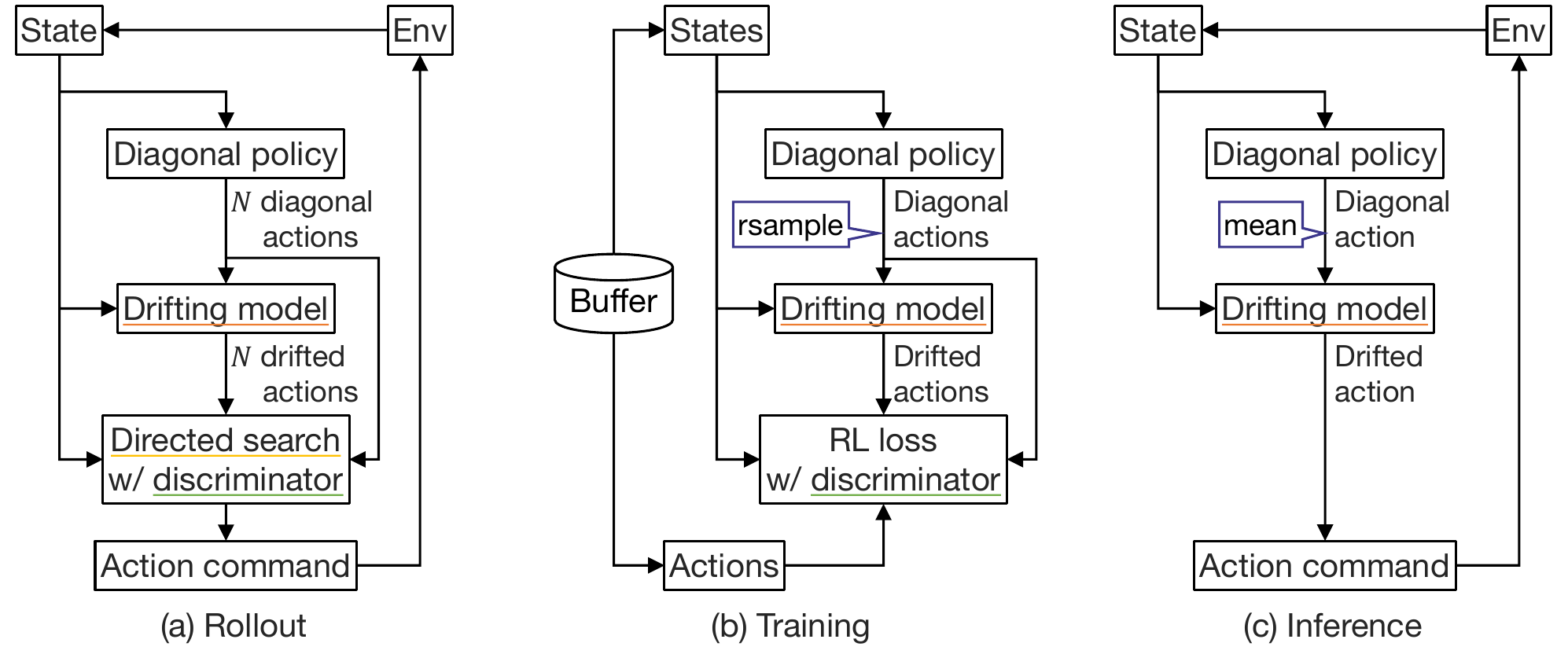}
    \caption{Flows with three techniques at three phases in the proposed algorithm}
    \label{fig:framework}
\end{figure}

This section will introduce three proposed techniques in order.
These techniques are utilized in three phases: (a) rollout for data collection; (b) training with the replayed data; and (c) inference and control using the trained policy (see Fig.~\ref{fig:framework}).
Note that the algorithms in this paper are limited to off-policy ones because the proposed method needs Q-gradient.
Furthermore, during inference, the mean of the diagonal policy, which serves as a noise source, is corrected by the drifting model; however, this differs from the mean of the drifted policy, which cannot be expressed in closed from and requires a Monte Carlo approximation.
To prioritize computational efficiency and high reproducibility, this paper employs the drifted mean.

\subsection{Drifting model as residuals}

The reason why replacing straightforward conventional policy model (e.g. diagonal Gaussian distribution) with the drifting model failed was that the model implicitly attempts to learn whole mappings from states to actions.
Therefore, as a first approach, a structure is designed so that the noise source for the drifting model is given a clear meaning, and the missing information is provided by the drifting model.
Specifically, by making the noise source a diagonal policy model, we enable the system to acquire a rough mapping between states and actions, while allowing the drifting model to learn action coordination as residuals.

To this end, let's first define the actions $a_{b,a} \in \mathcal{A}$ before and after the drifting as follows:
\begin{align}
    a_b \sim \pi_\phi(\cdot \mid s)
    \quad
    a_a = f_{\psi}(a_b; s)
\end{align}
where, $f_\psi: \mathcal{A} \times \mathcal{S} \mapsto \mathcal{A}$ denotes the drifting model parameterized by $\psi$.
In addition, by conditioning it over state $s$, it can represent the state-dependence of action coordination.
Note that in the actual implementation, $f_\psi$ is composed with a saturation function so that its output satisfies the action box constraints.

Although the corresponding policy, $\tilde{\pi}$, is only implicitly given (making it possible to sample but difficult to compute the likelihood), the RL algorithm with Q-gradient does not use the policy's likelihood, so it can be replaced with this model without issue.
However, as mentioned earlier, it is necessary to assign a explicit meaning to $a_b$ as well (i.e. action with inter-axis independence).
Therefore, this paper considers the following optimization problem.
\begin{align}
    \begin{split}
        \theta^\ast &= \arg \min_{\theta} \mathbb{E}_{(s_i, a_i, s^\prime_i, r_i) \sim D}\left[ (r_i + \gamma \mathrm{sg}(Q_\theta(s^\prime_i, a_a^\prime)) - Q_\theta(s_i, a_i) )^2 \right]
        \\
        \phi^\ast, \psi^\ast &= \arg \min_{\phi,\psi} \mathbb{E}_{(s_i) \sim D, m_i \sim \mathcal{B}(p)}\left[ - Q_\theta(s_i, \tilde{a}) \right]
    \end{split}
\end{align}
s.t.
\begin{align*}
    a_a^\prime &= f_{\psi}(a_b^\prime; s^\prime_i) \quad a_b^\prime \sim \pi_\phi(\cdot \mid s^\prime_i)
    \nonumber \\
    \tilde{a} &= m_i f_{\psi}(a_b; s^\prime_i) + (1-m_i) a_b \quad a_b \sim \pi_\phi(\cdot \mid s_i)
\end{align*}
where, $\mathcal{B}(p)$ denotes Bernoulli distribution for masking with $p \in [0, 1]$ the probability as a hyperparameter.
In other words, while optimizing $Q$ based on the policy with drifting, policy improvements are performed both before and after drifting to ensure that the action at stage $a_b$ is reasonably optimal.
Furthermore, as is commonly done in multi-agent RL \citep{wen2022multi}, by determining $a_a$ as an implicit random variable based on the opinion of $a_b$, we can correct action with residuals, which are expected to be action coordination.
Note again that this paper adds terms related to policy entropy to this base RL algorithm; however, since it is difficult to calculate the likelihood after drifting, the entropy of the pre-drifting policy is added, preventing collapse of the pre-drifting policy.

\subsection{Discriminator as regularization}

Based on the above implementation, it is expected that the drifting model will be able to structurally represent action coordination.
However, policy improvements via gradient-based methods does not guarantee this.
Therefore, a regularization term is introduced to promote role differentiation in the drifting model and the noise source preceding it.

Specifically, a discriminator that distinguishes between the actions $a_{b,a}$ before and after drifting is trained by adversarial learning \citep{goodfellow2020generative}.
In other words, $a_b$ (and its source, $\pi$) is regularized so that the discriminator misidentifies it as $a_a$, while $a_a$ (and the drifting model $f_\psi$) is regularized so that the discriminator can easily distinguish it from $a_b$.
As a result, $a_b$ should align as closely as possible with $a_a$ within the constraints of inter-axis independence of actions, while $a_a$ should be encouraged to acquire unique information (i.e. action coordination) to distinguish itself from $a_b$.
Such a regularization term is given below.
\begin{align}
    \mathcal{L}_d(\phi, \psi, \eta) &= -\beta \mathbb{E}_{(s_i) \sim D} [(1 - \rho) \ln d_\eta(1 \mid s_i, a_a) + \rho \ln d_\eta(0 \mid s_i, \mathrm{GRL}(a_b))]
\end{align}
s.t.
\begin{align*}
    \begin{split}
        \rho = \mathrm{sg}\left\{ \frac{d_\eta(1 \mid s_i, a_a)}{d_\eta(1 \mid s_i, a_a) + d_\eta(0 \mid s_i, a_b)} \right\}
        \quad
        a_b \sim \pi_\phi(\cdot \mid s_i)
        \quad
        a_a = f_\psi(a_b; s_i)
    \end{split}
\end{align*}
where, $\beta \geq 0$ is the regularization gain and is a hyperparameter.
To simplify the implementation of adversarial learning, a gradient reversal layer $\mathrm{GRL}(\cdot)$, which reverse teh sign of the gradient \citep{ganin2016domain}, is introduced.
Note that $\rho$ is a heuristic designed to balance between the two terms and mitigate learning that is biased toward one side.
We solve this regularization term simultaneously with the original optimization problem.

\subsection{Directed search with two directions}

For the discriminator-based regularization to work effectively, the data distribution used for training is crucial.
On the one hand, data that is difficult to classify and leads to misclassification is considered more valuable.
On the other hand, in regions where classification is generally accurate, value would remain in new regions that can be reached through action coordination.
Since these two aspects represent a kind of exploration-exploitation dilemma \citep{thrun1992efficient}, it is difficult to pursue them simultaneously.
Therefore, a trick, which chooses one of these two directions and prioritizes the collection of high-value data, is introduced.
Note that this directed search is available as the target RL algorithm is off-policy.

At first, given the current state $s$, we sample $N$ candidates.
\begin{align}
    a_{b,i} \sim \pi_\phi(\cdot \mid s) \quad a_{a,i} = f_\psi(\{a_{b,i}\}_{i=1}^N; s) \quad (i=1,\ldots,N)
\end{align}
The probabilities of the discriminator corresponding to these are given as follows:
\begin{align}
    d_{b,i} = d_\eta(1 \mid s, a_{b,i}) \quad d_{a,i} = d_\eta(1 \mid s, a_{a,i}) \quad (i=1,\ldots,N)
\end{align}

If the region represented by $s$ is already trained, $\{d_{a,i}\}_{i=1}^N \simeq 1$ and $\{d_{b,i}\}_{i=1}^N \simeq 0$ are expected; otherwise, it suggests that the learning is insufficient.
Therefore, the classification results to be used in the subsequent candidate selection are selected probabilistically.
\begin{align}
    \{d_i\}_{i=1}^N &= m \{d_{a,i}\}_{i=1}^N + (1 - m) \{d_{b,i}\}_{i=1}^N
    \\
    m &\sim \mathcal{B}\left( \frac{1}{2N}\left\{ \sum_{i=1}^N d_{a,i} + (1 - d_{b,i}) \right\} \right)
    \nonumber
\end{align}

Furthermore, the discriminator represents the likelihood ratio of the two probability distributions being compared as $\pi/\tilde{\pi} \simeq (1-d)/d$.
Since the negative logarithm of this corresponds to the mutual information generated by drifting, it is utilized as the objective function to be maximized for selecting one action from $N$ candidates.
This implies selecting values of $d_i$ that are close to $1$; as a result, we can expect either misclassification in under-learned regions or strong action coordination in over-learned regions.
In this paper, E-BoN sampling with $\alpha \in \mathbb{R}$ as the hyperparameter \citep{kobayashi2026flexible}, which is an extension of standard best-of-N sampling \citep{stiennon2020learning}, is employed as the selection mechanism.
Following the literature, we apply an equivalent transformation to enable normalization by the mean of the objective function, and the final action is selected as follows:
\begin{align}
    a &\sim \mathcal{C}(\{a_{a,i}\}_{i=1}^N \mid \{P_i\}_{i=1}^N)
    \\
    \{P_i\}_{i=1}^N &= \mathrm{Entmax}_\alpha\left( \left\{ \frac{J_i}{N^{-1}\sum_{j=1}^N J_j} \right\}_{i=1}^N \right)
    \nonumber \\
    J_i &= - \ln\left( \frac{1 - d_i}{d_i} \right) + \exp\left( \frac{1 - d_i}{d_i} \right) - 1 \quad (i=1,\ldots,N)
    \nonumber
\end{align}
where, $\mathcal{C}(\cdot \mid p)$ denotes the categorical distribution with probability $p$.

\section{Experiments}

\subsection{Setup}

To validate the proposed method, its learning performance is evaluated using three locomotion tasks provided by dm\_control \citep{tunyasuvunakool2020dm_control}, which is wrapped by Shimmy \citep{jun2023shimmy}.
The tasks are \textit{quadruped-walk-v0} (Quadruped; $|\mathcal{A}|=12$), \textit{humanoid-walk-v0} (Humanoid; $|\mathcal{A}|=21$), and \textit{dog-walk-v0} (Dog; $|\mathcal{A}|=39$).
The number of steps per episode is limited to 500.
For statistical evaluation, training is conducted under identical conditions using 15 random seeds.

The baselines are the standard diagonal policy (Vanilla) and a simple replacement of the policy model with the drifting model (Naive), in which the noise source in the drifting model is not explicitly learned as a diagonal policy.
The proposed method is compared with them by gradually adding the three tricks (D1, D2, and D3).
Note that each of the three proposals has its own hyperparameters; a simple sweep of them is included in Appendix.
Details regarding the implementation are also included in Appendix.

\subsection{Results}

\begin{figure}[!t]
    \centering
    \includegraphics[keepaspectratio=true,width=0.96\linewidth]{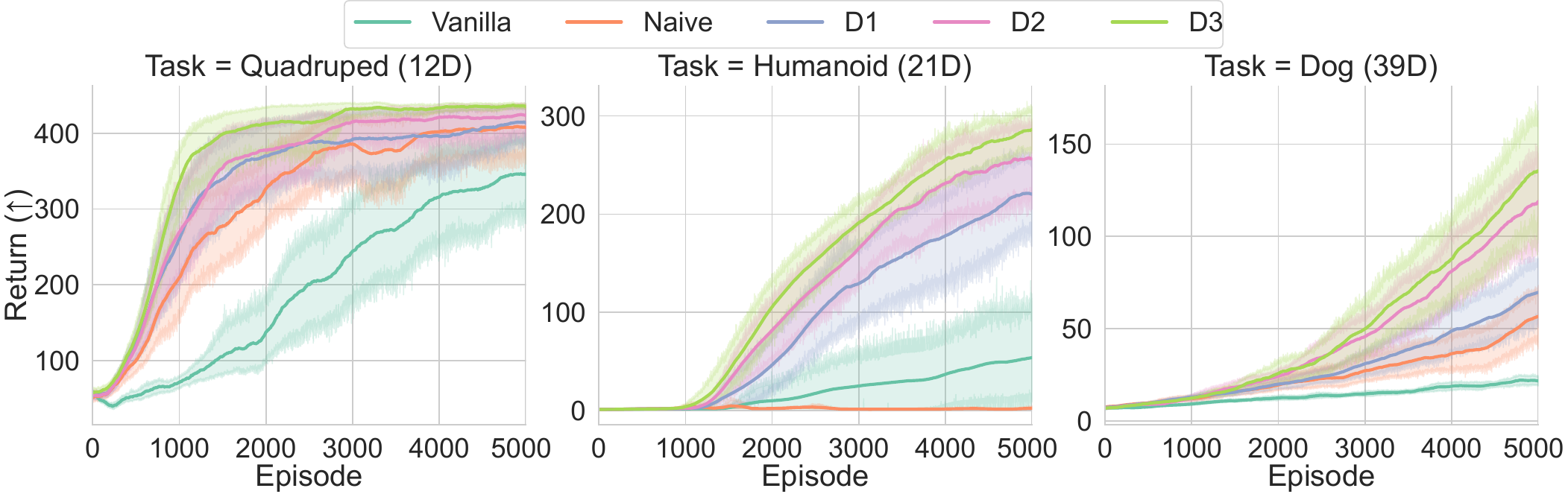}
    \caption{Learning curves of episodic returns}
    \label{fig:learn_curve}
\end{figure}

\begin{figure}[!t]
    \centering
    \includegraphics[keepaspectratio=true,width=0.84\linewidth]{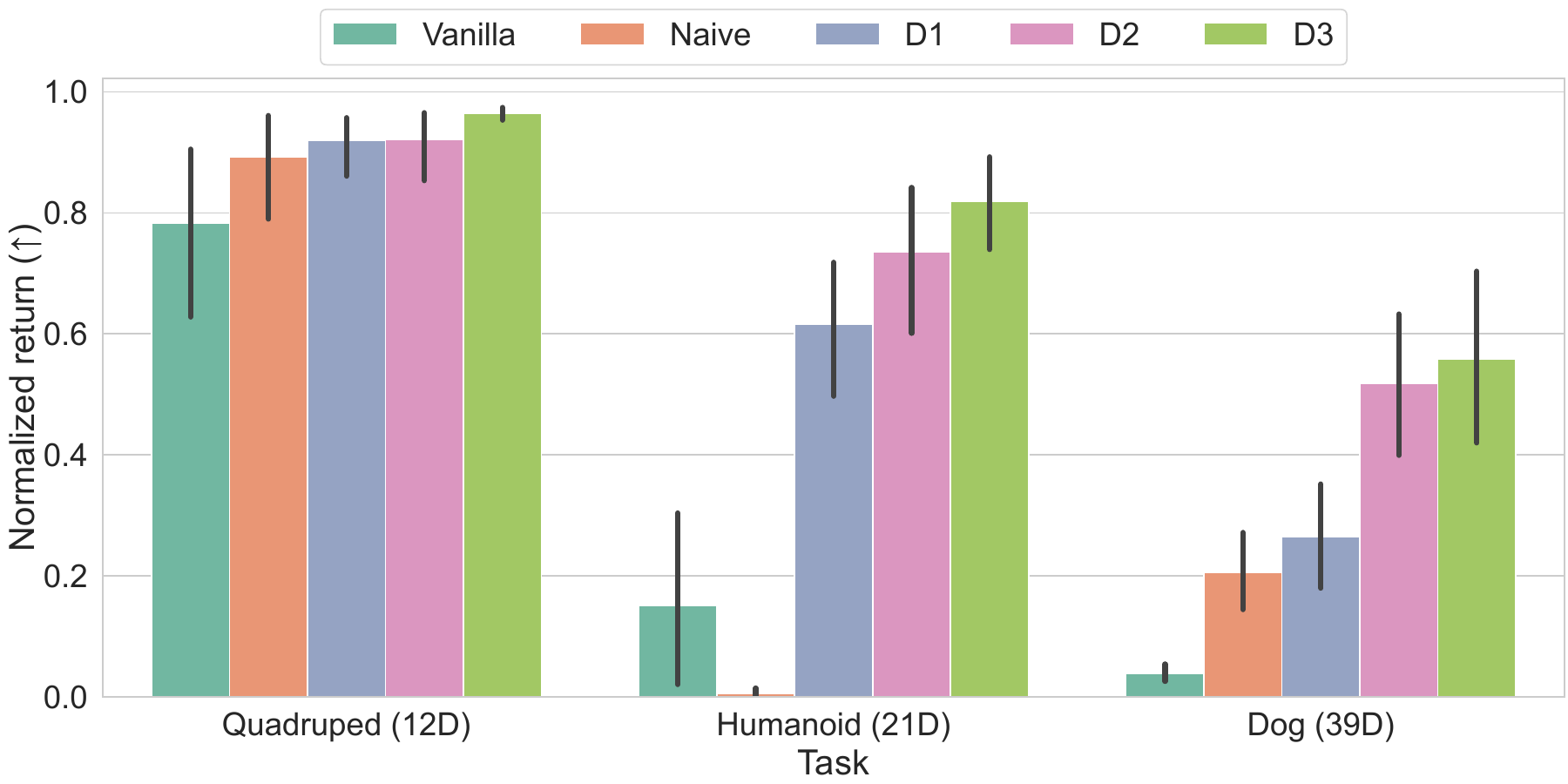}
    \caption{Test results with trained policies}
    \label{fig:test_return}
\end{figure}

First, the learning curves for episodic returns are depicted in Fig.~\ref{fig:learn_curve} for evaluating the sample efficiency.
In addition, the trained policies were tested 100 times, and the interquartile of them were normalized for evaluating the final control performance, as shown in Fig.~\ref{fig:test_return}.
Under all tasks, Vanilla learned more slowly than the others or failed to achieve satisfactory improvements in control performance, indicating that exploration without action coordination is difficult.
Furthermore, although Naive is capable of acquiring action coordination, its learning progress was similarly poor, suggesting that it is difficult to passively acquire action coordination through Q-gradient.
On the other hand, both the sample efficiency and control performance improved steadily as the three proposals were added in stages.
In particular, D1, which incorporates a diagonal policy to the noise source, enabled learning even on tasks where the baselines failed to learn, suggesting its effectiveness.

As a remark, another possibility not explored in this paper (namely, pre-training the diagonal policy, then freezing it (or reducing its learning rate) and using the drifting model as an adapter for post-training) is expected to fail.
That is, based on the Vanilla results, the diagonal policy cannot be made to function during pre-training.
Then, based on the Naive results, since the noise source does not play a role during post-training, the system would likely fail to discover action coordination.

\begin{figure}[!t]
    \centering
    \includegraphics[keepaspectratio=true,width=0.96\linewidth]{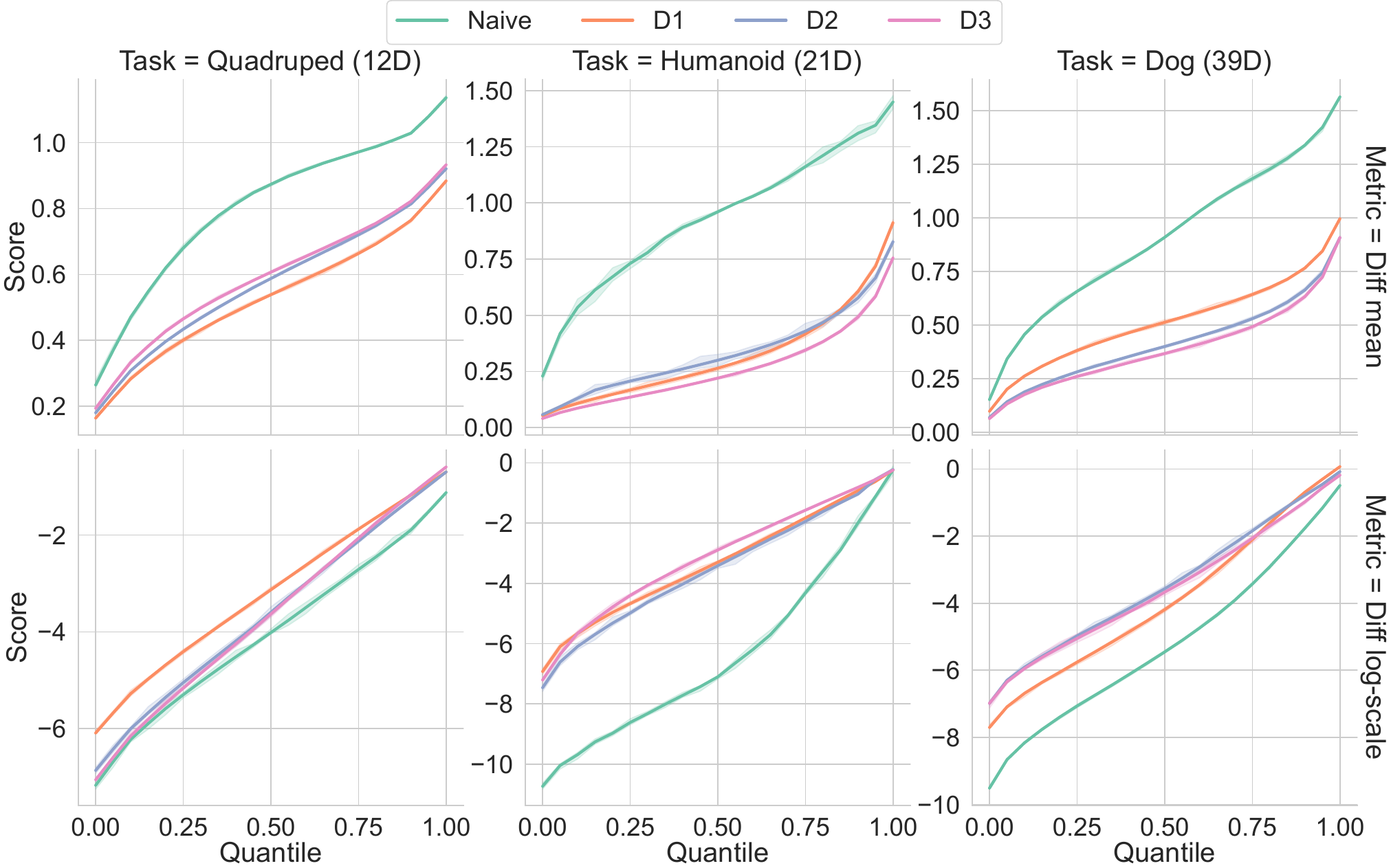}
    \caption{Analysis of drifting effects}
    \label{fig:policy_drift}
\end{figure}

Next, the extent to which the policy changed before and after drifting is analyzed.
The first metric is the absolute value of the shift in the distribution location, and the second is the logarithmic change in the distribution scale.
The trained policies of them were calculated for each action axis at each time step, and plotted with their quantile as the horizontal axis (see Fig.~\ref{fig:policy_drift}).
Since the noise source in Naive was meaningless, the location shifted significantly, and the scale also became small in most cases.
In contrast, the proposed method generated differences only where necessary.
In particular, the difference does not necessarily widen as more of the three techniques were applied; depending on the task, the difference is suppressed.
This result suggests that the drifting model corrected the residuals only to the extent necessary in conjunction with the diagonal actions.
However, while this is acceptable compared to Naive, the scale of the distribution mostly became smaller due to drifting, and there remains a possibility that the exploration is limited.

\section{Conclusion}

This paper proposes a novel off-policy RL algorithm with Q-gradient, capable of efficiently and stably solving RL problems with high-dimensional action spaces.
To scalably extract action coordination, which frequently occurs in high-dimensional action spaces, the drifting model was introduced, following three techniques to ensure it effectively extracts action coordination.
Specifically, (a) the drifting model was trained as residuals from the diagonal policy; (b) the discriminator was designed to only extract the fundamental differences between the policy before and after drifting; and (c) the directed search was introduced based on training progress of the discriminator.
As a result, the proposed method succeeded in significantly improving learning performance on the three locomotion tasks with high-dimensional action spaces.

However, although the experimental results confirmed that the proposed method successfully correct the diagonal actions as action coordination, there is concern that the exploration may be insufficient as the scale of the drifted policy decreased.
It would be allowable if it were determined uniquely through coordination with other action axes, but this might cause optimization failure especially if the decision is incorrect.
For example, such a premature restriction of exploration space could likely be resolved by maximizing the entropy for the drifted policy, or by introducing a strategy that occasionally uses the diagonal policy for data collection.
Updating the proposed algorithm based on such new approaches will make end-to-end learning for robotic systems with higher-dimensional action spaces (e.g. musculoskeletal robots), a key challenge for the future.

\subsection*{Generative AI Use}

Generative AI tools were used for editing.
The author reviewed and verified all AI-assisted content and take responsibility for the final content of the paper.

\acks{
This work was supported by JST CRONOS Grant Number JPMJCS24K6 and JSPS KAKENHI Grant Number JP26K03011.
}

\appendix
\section{Configuration details for experiments}

The benchmark tasks in this paper are selected from dm\_control \citep{tunyasuvunakool2020dm_control} wrapped by Shimmy \citep{jun2023shimmy}.
The maximum time steps per episode is basically limited to 500.
Evaluation is based on the episodic return (i.e. higher is better).
For statistical evaluation, training is conducted under identical conditions using 15 random seeds.

For the base RL algorithm, this paper follows the literature \citep{kobayashi2026flexible}.
In addition, the drifting model and discriminator are approximated by MLP with two layers and 100 neurons for each basically.
The drifting model is squashed so that its output corresponds to the action dimension (i.e. $[-1, 1]^{|\mathcal{A}|}$), and the discriminator output is passed through a sigmoid function to ensure $[0, 1]$, thereby outputting Bernoulli distribution.
However, to stabilize the computation, these output saturation functions are based on the method described in the literature \citep{barron2021squareplus}.
Furthermore, since the Dog task has a larger state and action spaces than the other tasks and was prone to unstable learning, the number of neurons in all models were doubled to 200 and set the learning rate to $10^{-5}$, which is $0.1$ times of the original value.
The impact of the proposed method's hyperparameters is investigated in the following experiments and the best setting from among them is selected.

\section{Sweeping three hyperparameters}

\begin{figure}[h]
    \subfigure[$p$]{
        \includegraphics[width=0.3175\linewidth]{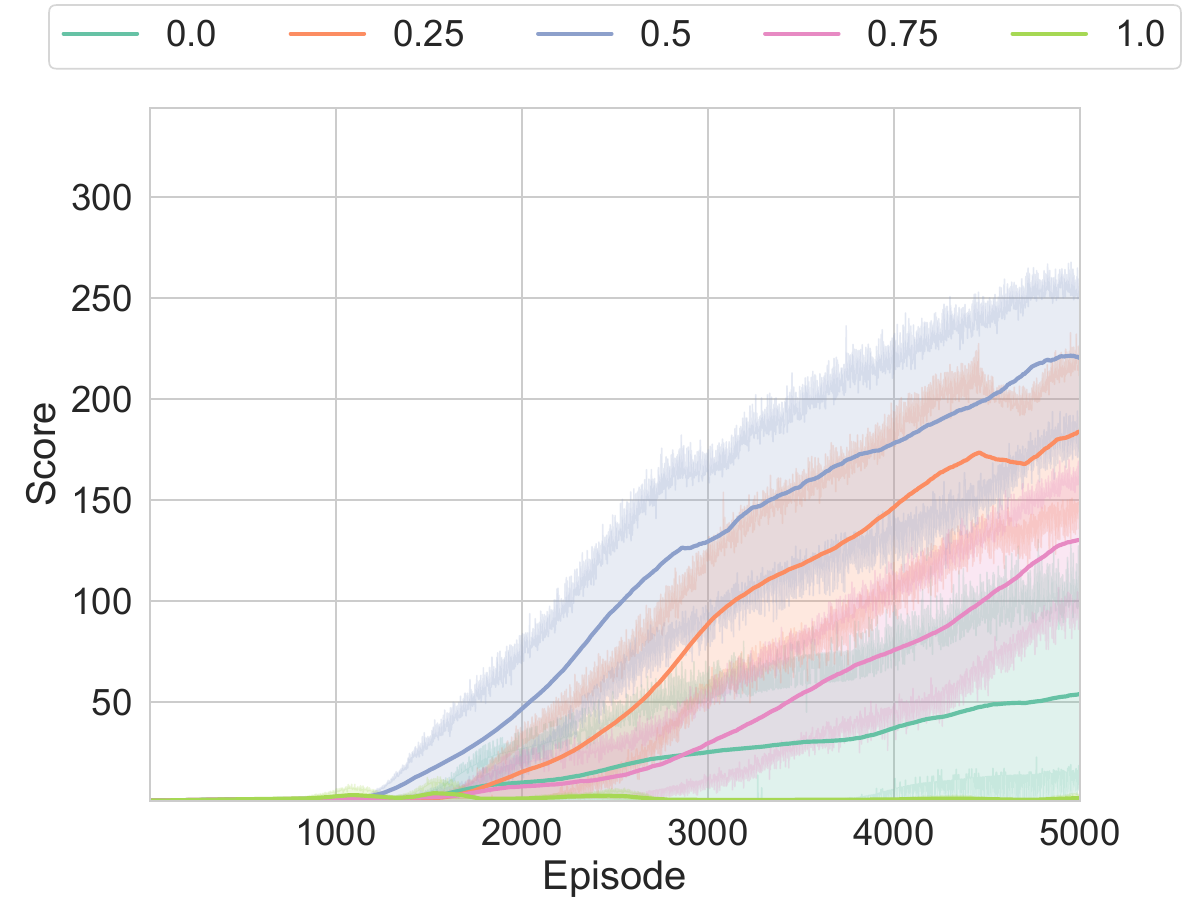}
        }
    \subfigure[$\beta$]{
        \includegraphics[width=0.3175\linewidth]{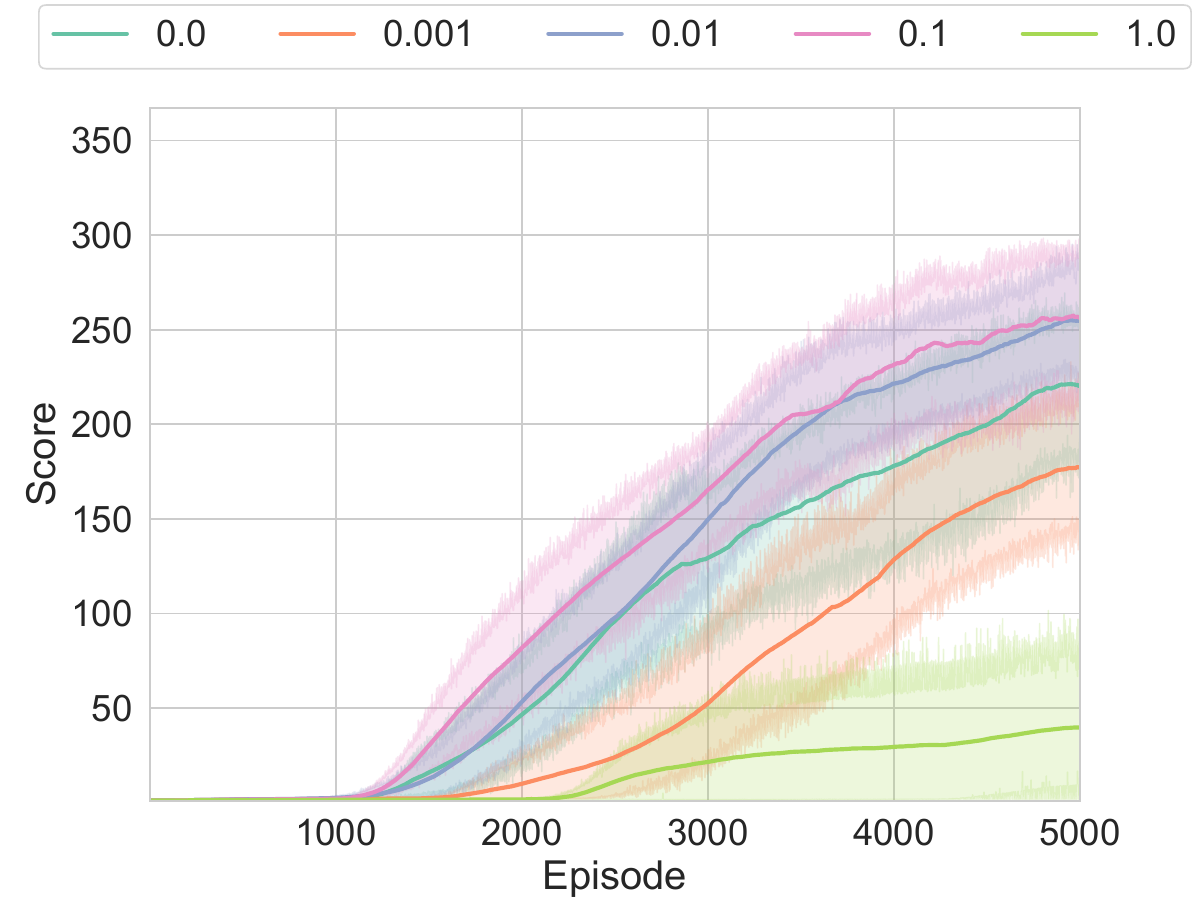}
        }
    \subfigure[$\sigma(\alpha-1)$]{
        \includegraphics[width=0.3175\linewidth]{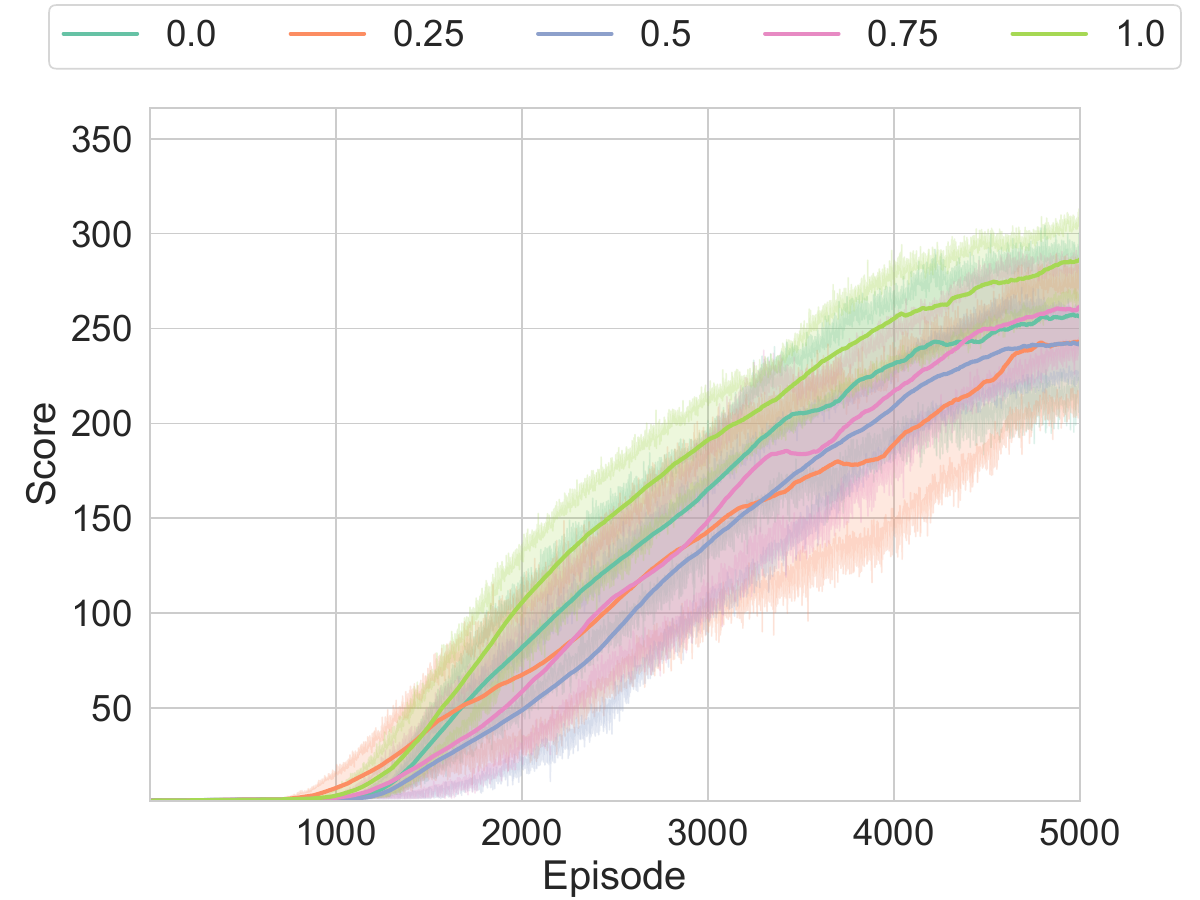}
        }
    \caption{Evaluation of the proposed approximation for entmax}
    \label{fig:ablation}
\end{figure}

In the proposed method, each of the three tricks has one hyperparameter.
The impact of these hyperparameters under the conditions described above is investigated, using \textit{humanoid-walk-v0} (Humanoid; $|\mathcal{A}|=21$).
In this investigation, each hyperparameter is swept at five points, spaced at regular intervals.
Due to the structure of the method, the tricks must be added sequentially; therefore, when investigating the second and third tricks, the best hyperparameters for the tricks applied up to that point are employed.

First, when training the drifting model as residuals, the probability $p \in [0, 1]$ of Bernoulli distribution is given by $p=\{0.0, 0.25, 0.5, 0.75, 1.0\}$.
Note that $p=0.0$ regards the Vanilla case as it never trains the drifting model; so only in this case, the drifting model is excluded completely.
Furthermore, $p=1.0$ corresponds to the Naive case as it never binds the noise source to the diagonal policy.
Next, for (b) the strength of regularization by the discriminator, $\beta \geq 0$, $\beta = \{0.0, 0.001, 0.01, 0.1, 1.0\}$ is set.
$\beta=0.0$ corresponds to not using the discriminator and is equal to the best result from (a).
Finally, for (c) E-BoN sampling \citep{kobayashi2026flexible} during directed search with $\alpha \in \mathbb{R}$, since dividing this into equal intervals is counterintuitive and difficult, it is normalized using $\sigma(\cdot): \mathbb{R} \mapsto [0, 1]$ and then adjusted so that the softmax case with $\alpha=1$ is its center.
That is, $\sigma(\alpha-1) = \{0.0, 0.25, 0.5, 0.75, 1.0\}$ is given.
Here, $\sigma(\alpha-1) = 0.0$ corresponds to a completely random selection from $N$ candidates, so it is equal to the best result from (b).

Fig.~\ref{fig:ablation} shows the training results for the settings described above.
First, regarding the effect of $p$, performance declined when $p$ is set too high, likely because the noise source struggles to function as a diagonal policy, so $p \leq 0.5$ can be considered appropriate.
In this paper, $p=0.5$ is regarded as the best value, since it yields the highest performance and is intuitively natural.
Next, regarding $\beta$, adding only a slight amount of regularization with $\beta=0.001$ did not work well, while excessively strong regularization with $\beta=1.0$ hindered the original RL progress.
A moderate strength in the range of $\beta = [0.01, 0.1]$ appears ideal, and in this paper, $\beta = 0.1$, which resulted in faster learning, is given to be the best choice.
Finally, regarding $\alpha$, while its impact is smaller compared to the above two tricks, we can find a performance improvement with $\alpha = \infty$, which corresponds to the argmax function without probabilistic selection.
For now, $\alpha=\infty$ is set to be the best in this paper.
However, with this trick, it might be better to fix that value and introduce the other hyperparameter for other factors, such as which of the two directions to be selected more frequently.

\bibliography{main}

@book{sutton2018reinforcement,
  title={Reinforcement learning: An introduction},
  author={Sutton, Richard S and Barto, Andrew G},
  year={2018},
  publisher={MIT press}
}

@inproceedings{rudin2022learning,
  title={Learning to walk in minutes using massively parallel deep reinforcement learning},
  author={Rudin, Nikita and Hoeller, David and Reist, Philipp and Hutter, Marco},
  booktitle={Conference on robot learning},
  pages={91--100},
  year={2022},
  organization={PMLR}
}

@article{radosavovic2024real,
  title={Real-world humanoid locomotion with reinforcement learning},
  author={Radosavovic, Ilija and Xiao, Tete and Zhang, Bike and Darrell, Trevor and Malik, Jitendra and Sreenath, Koushil},
  journal={Science Robotics},
  volume={9},
  number={89},
  pages={eadi9579},
  year={2024},
  publisher={American Association for the Advancement of Science}
}

@inproceedings{van2024revisiting,
  title={Revisiting reward design and evaluation for robust humanoid standing and walking},
  author={Van Marum, Bart and Shrestha, Aayam and Duan, Helei and Dugar, Pranay and Dao, Jeremy and Fern, Alan},
  booktitle={IEEE/RSJ International Conference on Intelligent Robots and Systems},
  pages={11256--11263},
  year={2024},
  organization={IEEE}
}

@article{liao2026beyondmimic,
  title={Beyondmimic: From motion tracking to versatile humanoid control via guided diffusion},
  author={Liao, Qiayuan and Truong, Takara E and Huang, Xiaoyu and Gao, Yuman and Tevet, Guy and Sreenath, Koushil and Liu, C Karen},
  journal={Science Robotics},
  volume={11},
  number={117},
  pages={eadx8924},
  year={2026},
  publisher={American Association for the Advancement of Science}
}

@article{liu2024masq,
  title={MASQ: multi-agent reinforcement learning for single quadruped robot locomotion},
  author={Liu, Qi and Guo, Jingxiang and Lin, Sixu and Ma, Shuaikang and Zhu, Jinxuan and Li, Yanjie},
  journal={arXiv preprint arXiv:2408.13759},
  year={2024}
}

@inproceedings{zhang2026falcon,
  title={FALCON: Learning Force-Adaptive Humanoid Loco-Manipulation},
  author={Zhang, Yuanhang and Yuan, Yifu and Gurunath, Prajwal and Gupta, Ishita and Omidshafiei, Shayegan and Agha-mohammadi, Ali-akbar and Vazquez-Chanlatte, Marcell and Pedersen, Liam and He, Tairan and Shi, Guanya},
  booktitle={Annual Learning for Dynamics and Control Conference},
  pages={265--281},
  year={2026},
  organization={PMLR}
}

@inproceedings{chou2017improving,
  title={Improving stochastic policy gradients in continuous control with deep reinforcement learning using the beta distribution},
  author={Chou, Po-Wei and Maturana, Daniel and Scherer, Sebastian},
  booktitle={International conference on machine learning},
  pages={834--843},
  year={2017},
  organization={PMLR}
}

@article{kobayashi2019student,
  title={Student-t policy in reinforcement learning to acquire global optimum of robot control},
  author={Kobayashi, Taisuke},
  journal={Applied Intelligence},
  volume={49},
  number={12},
  pages={4335--4347},
  year={2019},
  publisher={Springer}
}

@article{schwarke2025rsl,
  title={Rsl-rl: A learning library for robotics research},
  author={Schwarke, Clemens and Mittal, Mayank and Rudin, Nikita and Hoeller, David and Hutter, Marco},
  journal={arXiv preprint arXiv:2509.10771},
  year={2025}
}

@inproceedings{kim2026flashsac,
  title={Flashsac: Fast and stable off-policy reinforcement learning for high-dimensional robot control},
  author={Kim, Donghu and Lee, Youngdo and Park, Minho and Kim, Kinam and Nahendra, I Made Aswin and Seno, Takuma and Min, Sehee and Palenicek, Daniel and Vogt, Florian and Kragic, Danica and Peters, Jan and Choo, Jaegul and Lee, Hojoon},
  booktitle={Robotics science and systems},
  year={2026}
}

@article{chi2025diffusion,
  title={Diffusion policy: Visuomotor policy learning via action diffusion},
  author={Chi, Cheng and Xu, Zhenjia and Feng, Siyuan and Cousineau, Eric and Du, Yilun and Burchfiel, Benjamin and Tedrake, Russ and Song, Shuran},
  journal={The International Journal of Robotics Research},
  volume={44},
  number={10-11},
  pages={1684--1704},
  year={2025},
  publisher={Sage Publications, Sage UK: London, England}
}

@inproceedings{zhang2025flowpolicy,
  title={Flowpolicy: Enabling fast and robust 3d flow-based policy via consistency flow matching for robot manipulation},
  author={Zhang, Qinglun and Liu, Zhen and Fan, Haoqiang and Liu, Guanghui and Zeng, Bing and Liu, Shuaicheng},
  booktitle={AAAI Conference on Artificial Intelligence},
  volume={39},
  number={14},
  pages={14754--14762},
  year={2025}
}

@inproceedings{ren2025diffusion,
  title={Diffusion policy policy optimization},
  author={Ren, Allen and Lidard, Justin and Ankile, Lars and Simeonov, Anthony and Agrawal, Pulkit and Majumdar, Anirudha and Burchfiel, Benjamin and Dai, Hongkai and Simchowitz, Max},
  booktitle={International Conference on Learning Representations},
  volume={2025},
  pages={77288--77329},
  year={2025}
}

@inproceedings{mcallister2026flow,
  title={Flow matching policy gradients},
  author={McAllister, David and Ge, Songwei and Yi, Brent and Kim, Chung Min and Weber, Ethan and Choi, Hongsuk and Feng, Haiwen and Kanazawa, Angjoo},
  booktitle={International Conference on Learning Representations},
  volume={2026},
  pages={36352--36372},
  year={2026}
}

@inproceedings{deng2026drifting,
  title={Generative Modeling via Drifting},
  author={Deng, Mingyang and Li, He and Li, Tianhong and Du, Yilun and He, Kaiming},
  booktitle={Advances in Neural Information Processing Systems},
  year={2026}
}

@article{papamakarios2021normalizing,
  title={Normalizing flows for probabilistic modeling and inference},
  author={Papamakarios, George and Nalisnick, Eric and Rezende, Danilo Jimenez and Mohamed, Shakir and Lakshminarayanan, Balaji},
  journal={Journal of Machine Learning Research},
  volume={22},
  number={57},
  pages={1--64},
  year={2021}
}

@inproceedings{haarnoja2018soft,
  title={Soft actor-critic: Off-policy maximum entropy deep reinforcement learning with a stochastic actor},
  author={Haarnoja, Tuomas and Zhou, Aurick and Abbeel, Pieter and Levine, Sergey},
  booktitle={International conference on machine learning},
  pages={1861--1870},
  year={2018},
  organization={Pmlr}
}

@inproceedings{fujimoto2018addressing,
  title={Addressing function approximation error in actor-critic methods},
  author={Fujimoto, Scott and Hoof, Herke and Meger, David},
  booktitle={International conference on machine learning},
  pages={1587--1596},
  year={2018},
  organization={Pmlr}
}

@inproceedings{johannink2019residual,
  title={Residual reinforcement learning for robot control},
  author={Johannink, Tobias and Bahl, Shikhar and Nair, Ashvin and Luo, Jianlan and Kumar, Avinash and Loskyll, Matthias and Ojea, Juan Aparicio and Solowjow, Eugen and Levine, Sergey},
  booktitle={international conference on robotics and automation},
  pages={6023--6029},
  year={2019},
  organization={IEEE}
}

@article{wen2022multi,
  title={Multi-agent reinforcement learning is a sequence modeling problem},
  author={Wen, Muning and Kuba, Jakub and Lin, Runji and Zhang, Weinan and Wen, Ying and Wang, Jun and Yang, Yaodong},
  journal={Advances in Neural Information Processing Systems},
  volume={35},
  pages={16509--16521},
  year={2022}
}

@article{goodfellow2020generative,
  title={Generative adversarial networks},
  author={Goodfellow, Ian and Pouget-Abadie, Jean and Mirza, Mehdi and Xu, Bing and Warde-Farley, David and Ozair, Sherjil and Courville, Aaron and Bengio, Yoshua},
  journal={Communications of the ACM},
  volume={63},
  number={11},
  pages={139--144},
  year={2020},
  publisher={ACM, New York, NY, USA}
}

@article{kobayashi2024revisiting,
  title={Revisiting experience replayable conditions},
  author={Kobayashi, Taisuke},
  journal={Applied Intelligence},
  volume={54},
  number={19},
  pages={9381--9394},
  year={2024},
  publisher={Springer}
}

@article{stiennon2020learning,
  title={Learning to summarize with human feedback},
  author={Stiennon, Nisan and Ouyang, Long and Wu, Jeffrey and Ziegler, Daniel and Lowe, Ryan and Voss, Chelsea and Radford, Alec and Amodei, Dario and Christiano, Paul F},
  journal={Advances in neural information processing systems},
  volume={33},
  pages={3008--3021},
  year={2020}
}

@article{kobayashi2026flexible,
  title={Flexible Empowerment at Reasoning with Extended Best-of-N Sampling},
  author={Kobayashi, Taisuke},
  journal={arXiv preprint arXiv:2604.15614},
  year={2026}
}

@article{letizia2021discriminative,
  title={Discriminative mutual information estimators for channel capacity learning},
  author={Letizia, Nunzio A and Tonello, Andrea M},
  journal={arXiv preprint arXiv:2107.03084},
  year={2021}
}

@article{tunyasuvunakool2020dm_control,
  title={dm\_control: Software and tasks for continuous control},
  author={Tunyasuvunakool, Saran and Muldal, Alistair and Doron, Yotam and Liu, Siqi and Bohez, Steven and Merel, Josh and Erez, Tom and Lillicrap, Timothy and Heess, Nicolas and Tassa, Yuval},
  journal={Software Impacts},
  volume={6},
  pages={100022},
  year={2020},
  publisher={Elsevier}
}

@software{jun2023shimmy,
  author={Jun Jet Tai and Mark Towers and Elliot Tower},
  title={{Shimmy: Gymnasium and PettingZoo Wrappers for Commonly Used Environments}},
  year={2023},
  publisher={Zenodo},
  version={v1.1.0},
  doi={10.5281/zenodo.8140744},
}

@inproceedings{kingma2014auto,
  title={Auto-encoding variational bayes},
  author={Kingma, Diederik P and Welling, Max},
  booktitle={International Conference on Learning Representations},
  year={2014}
}

@article{ganin2016domain,
  title={Domain-adversarial training of neural networks},
  author={Ganin, Yaroslav and Ustinova, Evgeniya and Ajakan, Hana and Germain, Pascal and Larochelle, Hugo and Laviolette, Fran{\c{c}}ois and March, Mario and Lempitsky, Victor},
  journal={Journal of machine learning research},
  volume={17},
  number={59},
  pages={1--35},
  year={2016}
}

@book{thrun1992efficient,
  title={Efficient exploration in reinforcement learning},
  author={Thrun, Sebastian B},
  year={1992},
  publisher={Carnegie Mellon University}
}

@article{barron2021squareplus,
  title={Squareplus: A Softplus-Like Algebraic Rectifier},
  author={Barron, Jonathan T},
  journal={arXiv preprint arXiv:2112.11687},
  year={2021}
}

@article{han2021max,
  title={A max-min entropy framework for reinforcement learning},
  author={Han, Seungyul and Sung, Youngchul},
  journal={Advances in Neural Information Processing Systems},
  volume={34},
  pages={25732--25745},
  year={2021}
}

@inproceedings{nauman2025decoupled,
  title={Decoupled policy actor-critic: Bridging pessimism and risk awareness in reinforcement learning},
  author={Nauman, Michal and Cygan, Marek},
  booktitle={AAAI Conference on Artificial Intelligence},
  volume={39},
  number={18},
  pages={19633--19641},
  year={2025}
}

\end{document}